\documentclass[letterpaper, 10 pt, conference]{ieeeconf}  

\IEEEoverridecommandlockouts                              

\usepackage{graphicx}
\usepackage{float}
\usepackage{soul}
\graphicspath{ {./images/} } 

\usepackage{color}
\usepackage{amsmath}
\DeclareMathOperator*{\argmax}{argmax}
\usepackage{breqn}

\usepackage{algorithm}
\usepackage[noend]{algpseudocode}
\usepackage[usenames,dvipsnames]{xcolor}

\usepackage{balance}
\usepackage{amsfonts}

\usepackage{tikz}
\usetikzlibrary{matrix, positioning}

\title{\LARGE \bf
Optimal Sequential Stochastic Deployment of Multiple Passenger Robots
}

\author{Chris (Yu Hsuan) Lee, Graeme Best, Geoffrey A. Hollinger
\thanks{*Approved for public release; distribution is unlimited. This work was in part sponsored by DARPA under agreement \#HR00111820044 and Office of Naval Research Grant \mbox{N00014-17-1-2581}. Any opinions, findings and conclusions or recommendations expressed in this material are those of the authors and do not necessarily reflect those of the sponsor.}
\thanks{*A preliminary version of this work appeared as a workshop paper in~\cite{lee2020workshop}.}%
\thanks{*The authors are with the Collaborative Robotics and Intelligent Systems (CoRIS) Institute, Oregon State University, Corvallis OR, USA. 
	{\tt\small \{leeyuh,bestg,geoff.hollinger\}@oregonstate.edu}}%
}

\begin{document}

\maketitle
\thispagestyle{empty}
\pagestyle{empty}

\begin{abstract}
We present a new algorithm for deploying passenger robots in marsupial robot systems. A marsupial robot system consists of a carrier robot (e.g., a ground vehicle), which is highly capable and has a long mission duration, and at least one passenger robot (e.g., a short-duration aerial vehicle) transported by the carrier. We optimize the performance of passenger robot deployment by proposing an algorithm that reasons over uncertainty by exploiting information about the prior probability distribution of features of interest in the environment. Our algorithm is formulated as a solution to a sequential stochastic assignment problem (SSAP). The key feature of the algorithm is a recurrence relationship that defines a set of observation thresholds that are used to decide when to deploy passenger robots. Our algorithm computes the optimal policy in $O(NR)$ time, where $N$ is the number of deployment decision points and $R$ is the number of passenger robots to be deployed. We conducted drone deployment exploration experiments on real-world data from the DARPA Subterranean challenge to test the SSAP algorithm. Our results show that our deployment algorithm outperforms other competing algorithms, such as the classic secretary approach and baseline partitioning methods, and is comparable to an offline oracle algorithm. 
\end{abstract}

\section{Introduction}

Exploration of increasingly complex environments demands more flexible robotic capabilities. Developments in heterogeneous multi-robot systems have yielded carrier-passenger robot systems called ``marsupial robots", in which highly-capable carrier robots transport and deploy one or more low-capability passenger robots. These marsupial robots can tailor their complementary capabilities to the challenges of exploring complex environments~\cite{Murphy1999}. During exploration, these environments can contain multiple features of interest that the carrier robot may want to observe but are prohibitive or difficult to reach~\cite{tokekar2016sensor, moore2016nested}. In marine environments, a large ship may carry and deploy multiple heterogeneous robots to increase the rate of information gathering of features such as seafloor mines, adversarial vessels, and biological hotspots~\cite{das2015data, marques2015marsupial,kalaitzakis2020marsupial}. In the case of urban environments, a team of ground and aerial robots may seek to explore features like ledges, vertical shafts, stairways, or other urban features that ground robots cannot easily reach~\cite{rouvcek2019darpa}. During these missions, the carrier robot would ideally deploy the passenger robot at a location that can maximize exploration coverage or information gain and make use of the passenger robot's complementary capabilities (e.g., flying in 3D).  

\begin{figure}[tp]
    \centering
    \includegraphics[width=\linewidth]{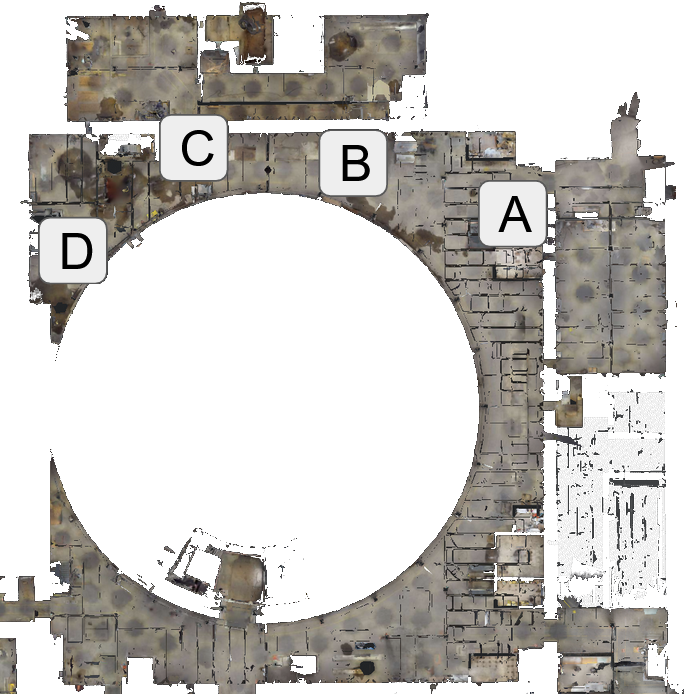}
    \caption{RGB point cloud map~\cite{matterport} of an urban environment from the DARPA Subterranean Challenge~\cite{allen_chung} with example deployment locations. Each decision point is revealed sequentially to the robot. Point A has high exploration potential since it leads to another room, but not as much as point C, which leads to multiple rooms and a stairwell only accessible to aerial passenger robots. However, points B and D are poor deployment locations since they lack nearby features to be explored by the passenger robots.}
    \vspace{-4ex}
    \label{fig:urban_environment}
\end{figure}

In order to optimize exploration efficiency, a marsupial robot system must decide when and where to deploy its passenger robots. In Fig.~\ref{fig:urban_environment}, an example of an urban environment is illustrated with sequential decision locations having different exploration potential. If a carrier robot deploys too early, it risks missing out on the potentially more valuable later decision points. Each of these sequential decision locations may have different reward values that are only revealed when they are observed along the robot's path. Due to the unknown nature of an unexplored environment, the carrier robot is required to make online deployment decisions while reasoning over the possible discovered deployment value at each potential deployment location. Furthermore, the value and number of these features are not known in advance. Multi-robot systems require coordination beyond naive partitioning of deployment between the passenger robots to ensure that the maximum expected number of features are captured~\cite{karapetyan2017efficient}. Prior works have explored marsupial robot coordination and planning~\cite{moore2016nested,wurm2013coordinating} but, to our knowledge, only deploy in a naive manner~\cite{couceiro2014marsupial}. To address these challenges, we develop an online passenger deployment strategy that reasons over the predicted future reward of deploying at each decision point.


We formulate the passenger robot deployment problem as a sequential stochastic assignment problem (SSAP)~\cite{derman1972sequential} where a set of passenger robots are assigned to deployment locations. The deployment algorithm exploits prior probabilistic information of the distribution of features throughout the environment. 
The key component of the algorithm is a recurrence relationship that defines a set of observation thresholds. These thresholds are used to decide when to deploy passenger robots by comparing to the current observation of deployment reward. 
The algorithm is guaranteed to find the optimal online deployment policy for problems with a known prior distribution with independent observations. Also, the algorithm is efficient in that it has $O(NR)$ runtime, where $N$ is the number of decision points and $R$ is the number of passenger robots. This computational complexity improves upon~\cite{derman1972sequential}, which had $O(N^2)$ complexity.

We conducted drone deployment exploration simulated experiments where a ground robot seeks to deploy drones in exploration-valuable deployment locations. The ground robot observes hard-to-reach frontier cells that are valuable for exploration by the aerial drone in each deployment location and must make a decision to deploy or not deploy the aerial drone. Data from the DARPA Subterranean (SubT) Challenge Urban circuit~\cite{teamExplorer} were used for the experiments. We show empirically that our algorithm is competitive with an offline oracle solution that has full access to the rewards in advance. Our algorithm also greatly outperforms comparable deployment algorithms, including a partitioned solution to the classic secretary problem~\cite{ferguson_2008}, and naive baseline methods. In the experiment, the algorithm is shown to be capable of choosing to deploy aerial drones at exploration-valuable deployment locations. On average, the SSAP algorithm performed within $96\%$ of the oracle in experiments with simulated data and performed within $90\%$ of the oracle on experiments with the real-world data.

Our contributions in this paper are: 
\begin{enumerate}
    \item A multi-robot passenger deployment problem formulated as a sequential stochastic assignment problem
    \item An informed passenger deployment strategy for this problem, which builds on~\cite{derman1972sequential}, that maximizes the expected sum of the deployment rewards.
\end{enumerate}

\section{Related Work}

Traditional heterogeneous robot systems have loosely-coupled motion constraints that are not physically constricted and have been coordinated with distributed planners~\cite{best2017path}. In marsupial robots, tightly-coupled constraints are physically imposed on the deployments of the passenger robots by the carrier robots. The increased complexity requires new planners that are able to handle tasks like coordination, deployment, retrieval, and manipulation. Some marsupial robot planners are able to organize and coordinate between multiple marsupial robotic systems to handle exploratory navigation actions~\cite{moore2016nested, wurm2013coordinating} but do not explicitly decide on optimal deployment locations for passenger robots. Other planners require known potential routes and deployment locations to decide optimal passenger robot deployment locations~\cite{mathew2015planning}. Another marsupial robot planner variant simplifies the deployment decision by deploying single passenger robots at fixed times such as the beginning of the mission~\cite{las2015path}. Finally, prior domain knowledge, such as marine flow fields~\cite{hansen2018autonomous}, has been used to inform multiple passenger robot planners to maximize information gain. However, these types of historical information field data may not be readily available for other domains such as urban environments. We present an algorithm that is capable of making multiple explicit passenger robot deployment decisions without a need to know potential routes. 

Our deployment problem is closely related to optimal stopping theory~\cite{ferguson_2008}, which considers problems that require deciding online the right time to carry out a particular action. A key example of an optimal stopping problem is the Classic Secretary problem, but it does not leverage prior information regarding future rewards~\cite{ferguson_2008}. However, the Cayley--Moser problem~\cite{moser1956problem} reasons over predicted future rewards by considering the prior probability distribution. This has been applied to robotics problems by Lindh{\'e} and Johansson~\cite{lindhe2013exploiting}, who consider the problem of when to communicate by utilizing a multi-path fading communication model to make predictions about future stopping decisions. Our method is a generalization of the Cayley--Moser algorithm, adapted for the context of the deployment problem. Additionally, we formulate a policy that considers all future deployments, rather than just the single next deployment. 

Optimal stopping variants have been generalized to multiple decision points. Das~et~al.~\cite{das2015data} apply multi-choice optimal stopping theory for AUVs to collect multiple plankton samples that minimize the cumulative regret of the samples. They explore two applicable algorithms, the multi-choice hiring algorithm~\cite{girdhar2009optimal} and the submodular secretary algorithm~\cite{bateni2010submodular}. Both of these algorithms seek to select the best observations and do not consider information such as the distribution of the observations. The sequential stochastic assignment problem (SSAP)~\cite{derman1972sequential} generalizes the Cayley--Moser algorithm and finds an optimal policy to maximize the expected sum of rewards from multiple assignments of agents to tasks. SSAP has been applied to other fields~\cite{shestak2012probabilistic} but, to our knowledge, has not been utilized in robotics. 
A closely related body of work focuses on multi-robot task assignment \cite{korsah2013comprehensive}; however, these problems generally assume that task values are known a priori and are therefore not directly applicable to online problems, such as ours.




\section{Problem Formulation}
\label{sec:problem_formulation}

We consider a marsupial robot system that must make online decisions regarding when to deploy its passenger robots. At each possible deployment location, the robot must consider if the reward gained from deployment is expected to be more favorable than continuing onwards and deploying at a later location. We formulate the multi-robot deployment decision as a sequential stochastic assignment problem (SSAP). Multiple deployment decisions are made based on sequentially revealed random variables. We formalize this problem as follows.

We consider an environment that contains a set of point features, which may represent points of interest that are ideal for additional exploration by passenger robots, but are ill-suited for the carrier robot. As the carrier robot moves through the environment, it must make decisions to deploy or not deploy a passenger robot. There are assumed to be a total of $N$ decision points, where $N$ is known to the robot. The carrier robot houses $R$ passenger robots of equal capability. At each decision point, the carrier robot may choose to deploy one passenger robot. Along the sequence of decision points, the independent observations of the number of features in the sensing area are denoted $(X_1,...,X_N)$. Furthermore, the algorithm relies on knowing a prior distribution of the random variable $X$, denoted as $f(x)$. Later in Sec.~\ref{sec:experiments}, we present an example where the features of interest correspond to the number of exploration frontier cells observed by the ground robot with a prior distribution $f(x)$ defined as a Poisson distribution.

At stage $j \in \{ 1,...,N \}$, the carrier robot reaches a decision point, and the outcomes of all random variables $(X_1,...,X_j)$, denoted $(x_1,...,x_j)$, are known to the robot. Along the path, the robot must make an irreversible decision at each deploy location to deploy or continue on to deploy later. If the carrier robot decides to deploy, it assigns one passenger robot to the deployment location and returns a reward of $x_j$. If the carrier robot decides to continue, no reward is claimed at this stage. This process continues for the $N$ stages. All passenger robots must be deployed by stage $N$, with the constraint of one deployment per stage.


We define the set of stage indices where passenger robots were deployed as $(d_1,...,d_R)$. The goal of the carrier robot is to maximize the expected sum of the reward returned from the deployment locations; i.e., find the optimal deployment sequence:
%
\begin{equation}
\label{eqn:objective}
    (d_1,...,d_R)^*=\argmax_{(d_1,...,d_R)}\mathbb{E} \Big[ \sum_{r=1}^{R}x_{d_r} \Big].
\end{equation}
Here, $x_{d_r}$ denotes the reward from passenger robot $r$ and the expectation is taken over the distribution $f(x)$.


\section{Online Passenger Deployment Algorithm}


We present the online passenger deployment algorithm, which finds the optimal deployment policy in linear time that scales with the number of passenger robots and deployment locations. The algorithm is computed via dynamic programming, using a technique that stems from sequential stochastic assignment~\cite{derman1972sequential}. We consider the general case of $R$ passenger robots to deploy. Finally, we provide an analysis of the optimality and runtime complexity of the algorithm. 


\subsection{Multi-Robot Deployment}
\label{sec:multi-robot-deployment}
Our deployment algorithm precomputes a set of thresholds, which are a function of the distribution of the observations and the number of remaining stages, $n$. The current observed value $x_j$ at stage $j$ is compared to the threshold values and informs the carrier robot whether or not to deploy now. At each stage $j$, we define a sequence of values $(p_1,...,p_n)$, such that $p_i=1$ for $n-r < i \le n$, representing the remaining $r$ passenger robots, and $p_i=0$ for $1 \le i \le n-r$. This can be thought of as assigning a $p_i$ at each stage to each $x_j$.

Specifically, the optimal policy for the passenger robot assignment is to assign $p_i$ to the observed deployment value $x_j$, if $x_j$ falls into an $i$th non-overlapping interval comprising the real line~\cite{derman1972sequential}. These non-overlapping intervals are separated by a set of thresholds, denoted as $a_{i,n}$. Each threshold $a_{i,n}$ is computed recursively and depends only on the prior distribution $f(x)$, as well as the number of remaining stages $n$. For stage $j$, where there are $n=N-j+1$ stages remaining, there are a set of $n+1$ thresholds, such that
\begin{equation}
\label{eqn:a_recurrence_line}
    -\infty=a_{0,n}\leq a_{1,n}\leq a_{2,n}\leq...\leq a_{n,n} = \infty.
\end{equation}
At stage $j$, the optimal choice is to assign $p_i$ if the realization $x_j$ of the random variable $X_j$ is contained in $(a_{i-1,n},a_{i,n}]$. For example, given $n=4$ remaining stages and $r=2$ robots to deploy, $p_3, p_4=1$ and $p_1, p_2=0$. If the observed value of $x_j$ is $a_{2,4} < x_j \leq a_{3,4}$, then the decision is to deploy and returns a reward of $p_3x_j = x_j$. At the next stage with $n=3$ and $r=1$, we only have $p_3=1$ and $p_1, p_2=0$. If the next observed value $x_j < a_{2,3}$, then there is no deploy action and returns a reward of $p_2\cdot x_j = 0$.

An assignment of $p_i=1$ to $x_j$ results in a deployment of the passenger robot whereas an assignment of $p_i=0$ to $x_j$ leads to a non-deployment action. There is no need to differentiate between interval containment below the deploy threshold $a_{i-R,n}$ since all of the $n-R$ assignments of $p_i=0$ result in the same non-deployment behavior. Thus, only $R$ thresholds, instead of $n+1$ as shown in~\eqref{eqn:a_recurrence_line}, need to be calculated for each stage $n$ beyond $n=R$, and this reduces the complexity from quadratic~\cite{derman1972sequential} to linear in $N$, as discussed later in Sec.~\ref{sec:analysis}.

The threshold $a_{i,n+1}$ is defined as the expected value, if there are $n$ stages remaining, of the quantity to which the $i$th smallest $p$ is assigned~\cite{derman1972sequential}. This formulates the recurrence relationship
\begin{align}
\label{eqn:drl_threshold_recurrence}
    a_{i,n+1} &= \Pr(x_{n}<a_{i-1,n}) a_{i-1,n} \nonumber \\
    &\quad \quad \quad
    + \Pr(a_{i-1,n} < x_{n} < a_{i,n}) \nonumber \\
    &\quad \quad \quad \quad \quad \quad \times \mathbb{E}(x_{n} | a_{i-1,n} < x_{n} < a_{i,n}) \nonumber \\
    &\quad \quad \quad
    + \Pr(x_{n} > a_{i,n}) a_{i,n} \\
    &= a_{i-1,n} \int_{-\infty}^{a_{i-1,n}}f(x)dx \nonumber
    +\int_{a_{i-1,n}}^{a_{i,n}}xf(x)dx \nonumber \\ \label{eqn:drl_threshold_recurrence2}
    &\quad \quad \quad 
    +a_{i,n} \int_{a_{i,n}}^{\infty}f(x)dx,
\end{align}
where $-\infty\cdot0 = 0$ and $\infty\cdot0 = 0$. In both \eqref{eqn:drl_threshold_recurrence} and \eqref{eqn:drl_threshold_recurrence2}, the second term is for the case where $x_{n}$ lies within the $i$th interval, and therefore $p_i$ receives the value of $x_{n}$. The first term is for the case where $x_{n}$ is below the $i$th interval, meaning that $x_{n}$ is assigned to a $p_k < p_i$, and thus $p_i$ is assigned in a later stage with an expected value of $a_{i-1,n}$. Similarly, the third term is for the case where $x_{n}$ is above the interval and $p_i$ has an expected future assignment of $a_{i,n}$. 

The recurrence process is illustrated in Fig.~\ref{fig:recurrence_calculation}, where the cells in the table can be computed from left to right by reusing the previous values, as indicated by the arrows. Furthermore, an outline of the algorithm is provided in Alg.~\ref{fig:pseudocode}. The outer loop iterates through the $N$ stages while the inner loop iterates through only $R$ thresholds. The algorithm's recurrence process has a computational complexity of $O(NR)$. In the $R=1$ single passenger robot deployment case, the threshold calculations are identical to the thresholds used in the Cayley--Moser optimal stopping problem~\cite{moser1956problem}.

\tikzset{ 
table/.style={
  matrix of nodes,
  row sep=-\pgflinewidth,
  column sep=-\pgflinewidth,
  nodes={rectangle,text width=3em,align=center},
  text depth=1.0ex,
  text height=1.5ex,
  nodes in empty cells
},
row 6/.style={nodes={fill=green!10}},
column 1/.style={nodes={fill=green!10}},
}

\begin{figure}[tp]
    \centering
    \begin{tikzpicture}
\matrix (mat) [table]
{
$4$ & - & - & - & $\infty$\\
$3$ & - & - & $\infty$ &$a_{3,4}$   \\
$2$ & - & $\infty$ &$a_{2,3}$ &$a_{2,4}$  \\
$1$ &$\infty$ &$a_{1,2}$ &$a_{1,3}$ & - \\
$0$ &$-\infty$& $-\infty$ &$-\infty$& $-\infty$ \\
 & $1$ & $2$ & $3$ & $4$ \\
};
\foreach \x in {1,...,5}
{
  \draw 
    ([xshift=-.5\pgflinewidth]mat-\x-1.south west) --   
    ([xshift=-.5\pgflinewidth]mat-\x-5.south east);
  }
\foreach \x in {1,...,4}
{
  \draw 
    ([yshift=.5\pgflinewidth]mat-1-\x.north east) -- 
    ([yshift=.5\pgflinewidth]mat-6-\x.south east);
}    
\draw (mat-6-1.south west)--(mat-6-1.north east);
\node[above right=2mm and 1.5mm of mat-6-1.south west] {$i$};
\node[below left=3mm and 2mm of mat-6-1.north east] {$n$};
\begin{scope}[shorten <=10pt,shorten >= 12pt]
\draw[->,line width=0.4mm]  (mat-3-3.center) -- (mat-3-4.center);
\draw[->,line width=0.4mm]  (mat-4-3.center) -- (mat-3-4.center);
\end{scope}

\end{tikzpicture}
    \caption{The method for the recurrence threshold calculation and terms relationship in \eqref{eqn:drl_threshold_recurrence}, for $R=2$ and $N=4$. $a_{i,n}$ is the threshold for the $i$th non-overlapping interval when there are $n$ stages left to go. The term $a_{2,3}$ is calculated from the $a_{i-1,n}$ term, which is $a_{1,2}$, and the $a_{i,n}$ term, which is $a_{2,2}=\infty$. $a_{1,4}$ does not need to be calculated since $R=2$.}
    \vspace{-1ex}
    \label{fig:recurrence_calculation}
\end{figure}

\begin{algorithm}[t]
\caption{\textit{SSAP} optimal policy threshold calculation for passenger robot deployment}
\label{fig:pseudocode}
\textbf{Input:} Number of stages: $N$, Number of robots: $R$, \\
               Prior distribution: $f(x)$ \\
\textbf{Output:} SSAP Thresholds: $a_{i,j}$
\begin{algorithmic}[1]
    \For{$j=0,1,2,\ldots,N$}
        \State $a_{j,j} = \infty$
        \For{$i=j-R,\ldots,j-1$}

        \State $a_{i,j} \leftarrow Eqn.~\eqref{eqn:drl_threshold_recurrence2}$
        \EndFor
    \EndFor
\end{algorithmic}
\end{algorithm}





\subsection{Analysis}
\label{sec:analysis}
The dynamic programming proceeds by iteratively solving optimal subproblems for $a_{i,n+1}$ using the recurrence relationship in \eqref{eqn:drl_threshold_recurrence2}, and thus computes the optimal set of thresholds $a_{i,n}, \forall i, n$. The full proof of the results that these subproblems are optimal, and that these thresholds lead to an optimal online assignment policy, is presented in~\cite{derman1972sequential}. The proof proceeds by induction, and relies on Hardy's Theorem~\cite{hardy1934inequalities}, which states that the optimal assignment between two sets with a sum-product objective is to pair the smallest values in each set, then the next smallest, and so on until the largest values are paired.

As illustrated in Fig.~\ref{fig:recurrence_calculation} and Alg.~\ref{fig:pseudocode}, there are $O(NR)$ sub-problems to be computed, where $N$ is the number of stages and $R$ is the number of passenger robots to deploy. The integral~\eqref{eqn:drl_threshold_recurrence2} is computed once per sub-problem, thus the computation time is $O(NRF)$, where $F$ is the time to compute~\eqref{eqn:drl_threshold_recurrence2}. 


\section{Experiments and Results}
\label{sec:experiments}
In order to evaluate our proposed SSAP passenger robot deployment algorithm, we performed drone deployment exploration experiments using simulated data and real-world data from the DARPA SubT challenge, against various other deployment strategies. 

\subsection{Comparison Methods}
Different deployment strategies were employed, each aiming to maximize the number of features captured during drone deployment, to test the efficacy of our SSAP deployment algorithm. These deployment strategies were adapted from optimal stopping algorithm variants to explicitly handle multiple robot deployments. As such, for these comparison methods, the carrier robot path is naively divided into $R$ equal partitions in the case of multi-robot deployment and treat each partition independently from each other. The deployment algorithms are described below:
\begin{itemize}
    \item \textit{Sequential Assignment (SSAP)}: Performs our method as described in Sec.~\ref{sec:multi-robot-deployment} and considers the entire path as a singular deployment, without partitions. The prior distributions are described in Sec.~\ref{sec:simulated_data_setup} and Sec.~\ref{sec:real-world-setup}.
    \item \textit{Oracle}: Selects the top $R$ deployment locations across all the partitions, with perfect knowledge of the world in advance.
    \item \textit{Partition Oracle}: Selects the best deployment location within each partition, with perfect knowledge of the world in advance.
    \item \textit{Partition Cayley--Moser}: Performs as described in Sec.~\ref{sec:multi-robot-deployment} when $R=1$, in each partition.
    \item \textit{Partition Classic Secretary}: An optimal stopping variant that only observes for the first $N/eR$, where $e$ is Euler's number, of decision points and then selects the next value that is higher than the max value observed in the observation phase~\cite{ferguson_2008}. The algorithm runs within each partition.
    \item \textit{Random}: Selects a decision point randomly within each partition.
\end{itemize}
The experiments were conducted on a standard desktop computer, with an i5-4690K CPU and 16~GB of RAM.


\subsection{Capturing Poisson-Distributed Features}
\subsubsection{Experimental Setup}
\label{sec:simulated_data_setup}

A carrier robot travels through a 2D simulated world containing features of interest distributed as a stationary Poisson point process. At decision point $j$, the carrier robot detects features within a circular sensing area centered at the current location. The number of features $x_j$ detected within a sensing area are modeled as a Poisson distribution with probability mass function
\begin{equation}
    \label{eqn:poisson}
    f(x) = \frac{\lambda^{x}e^{-\lambda}}{x!}, \text{for } x \in \{0,1,...\}.
\end{equation}
Here, the Poisson distribution $f(x)$ in~\eqref{eqn:poisson} is an example of a distribution used in the threshold calculations in~\eqref{eqn:drl_threshold_recurrence2}. The rate of this Poisson distribution, $\lambda$, is the expected number of features of interest in the robot sensing area. The \textit{SSAP} thresholds are calculated by substituting~\eqref{eqn:poisson} into~\eqref{eqn:drl_threshold_recurrence2}, yielding

\begin{align}
\label{eqn:pmf-drl-thresholds2}
a_{i,n+1} &= a_{i-1,n} \sum_{x=0}^{\lfloor a_{i-1,n} \rfloor} \frac{\lambda^{x}e^{-\lambda}}{x!}
+ \sum_{x=\lceil a_{i-1,n}\rceil}^{\lfloor a_{i,n} \rfloor} x  \frac{\lambda^{x}e^{-\lambda}}{x!} \nonumber \\
& \quad \quad \quad \quad \quad \quad \quad+ a_{i,n} \left[ 1 - \sum_{x=0}^{\lfloor a_{i,n}\rfloor}  \frac{\lambda^{x}e^{-\lambda}}{x!} \right]
.
\end{align}

\subsubsection{Results}
\begin{figure}[tp]
    \centering
    \includegraphics[width=\linewidth, trim=8mm 22mm 8mm 31mm, clip]{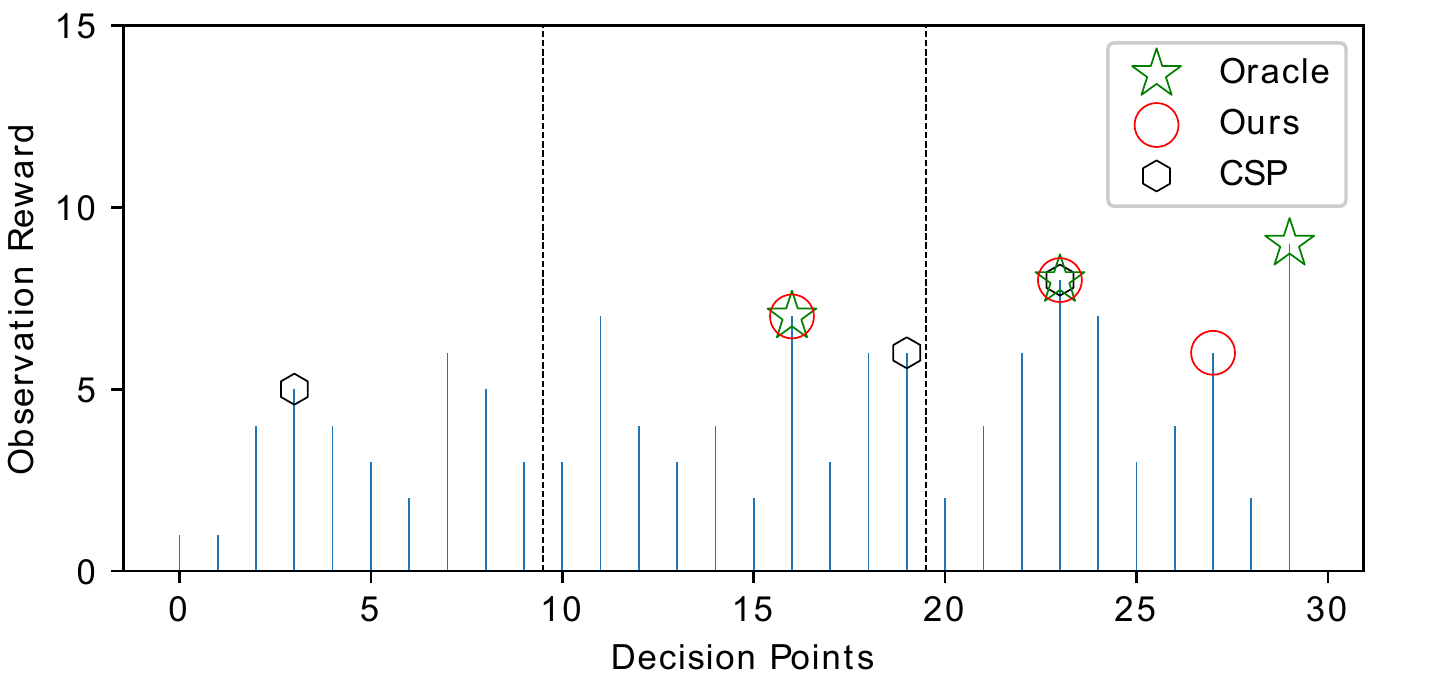}
    \caption{Illustrative example of the deployments for three passenger robots to be deployed over 30 decision points. Our method (reward: 21) performs similar to the \textit{Oracle} (reward: 24) that has full knowledge of the observation sequence in advance. The \textit{Classic Secretary} (reward: 19) algorithm performed worse, in part due to being constrained by having one deployment in each of the three partitions (dotted lines).}
    \vspace{-1ex}
    \label{fig:algorithm_example}
\end{figure}

An example of a trial of the algorithm selection process is illustrated in Fig.~\ref{fig:algorithm_example}. The \textit{Oracle} algorithm optimally selects the top $R=3$ rewards. \textit{SSAP} decided to deploy and select the reward in an optimal fashion in the first two decision cases. In comparison, the \textit{Classic Secretary} algorithm does not reason over possible future observations and locally selects a decision point in each partition. 

\begin{figure}[tp]
    \centering
    \includegraphics[width=\linewidth, trim=0mm 17mm 0mm 12mm, clip]{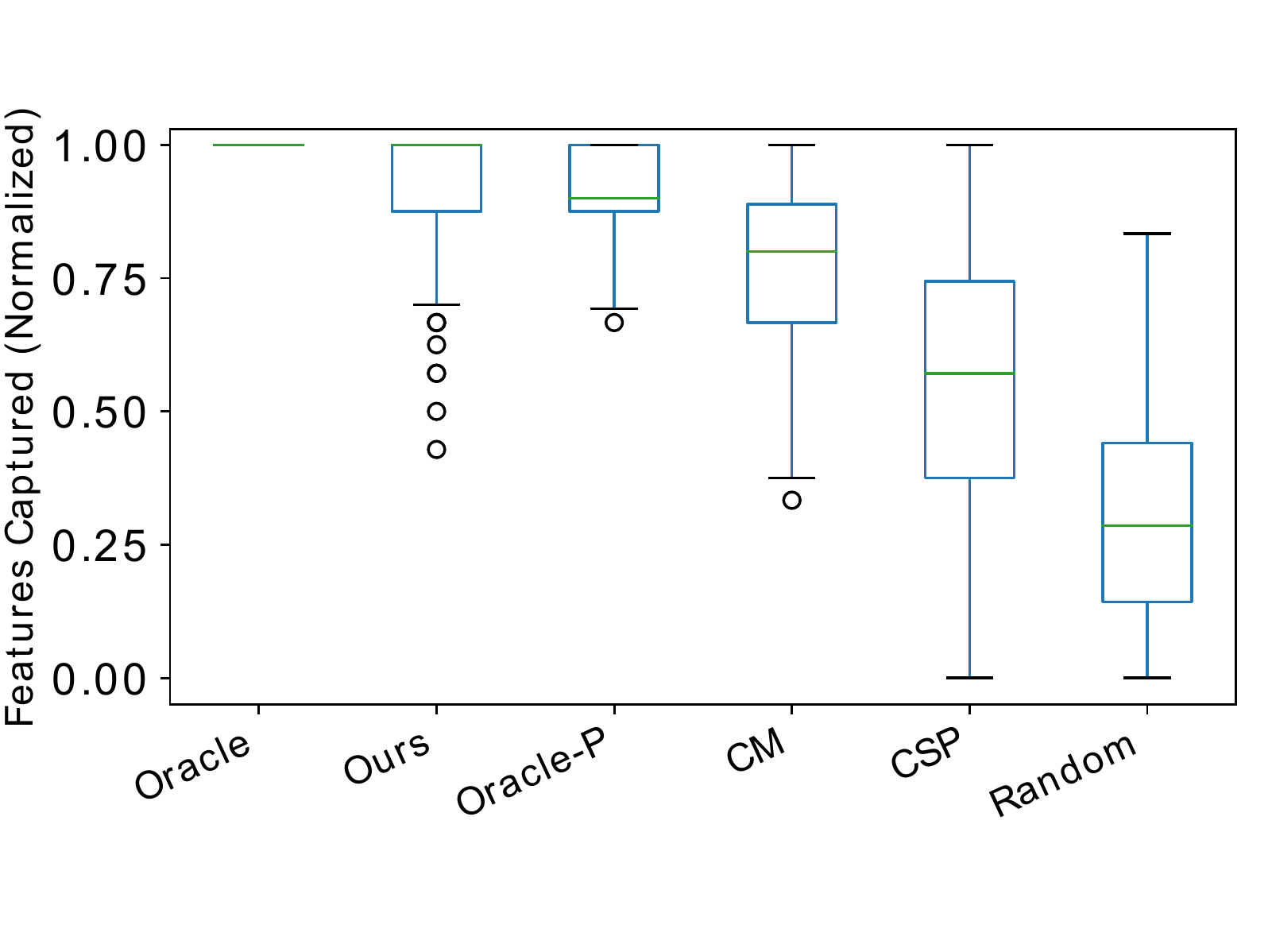}
    \caption{Algorithm comparisons with three passenger robots over 150 trials. \textit{Partition Oracle} is denoted as Oracle-P. CM is \textit{Cayley--Moser} and CSP is \textit{Classic Secretary}.}
    \vspace{-2ex}
    \label{fig:n_robot_results}
\end{figure}

The results of \textit{SSAP} through repeated trials are very encouraging. Trials with three passenger robots and $n=60$ stages are shown in Fig.~\ref{fig:n_robot_results}. During each trial, the algorithms' results are compared to the maximum possible number of features captured by the \textit{Oracle}. On average, \textit{SSAP} performed within $96\%$ of the \textit{Oracle}. Also, \textit{Partition Oracle} is not perfect like the \textit{Oracle} but performs similarly when compared to \textit{SSAP}.

The partitioned \textit{Cayley--Moser} algorithm has a lower performance than \textit{SSAP} since the calculated thresholds are only local to each partition and partitioned algorithms are constrained to one deployment location per partition. \textit{SSAP} is able to account for the expected future values over the total number of stages, whereas the partitioned \textit{Cayley--Moser} locally calculated the thresholds only for the number of stages in a partition. 

\begin{figure}[t]
    \centering
    \includegraphics[width=\linewidth, trim=0mm 15mm 0mm 24mm, clip]{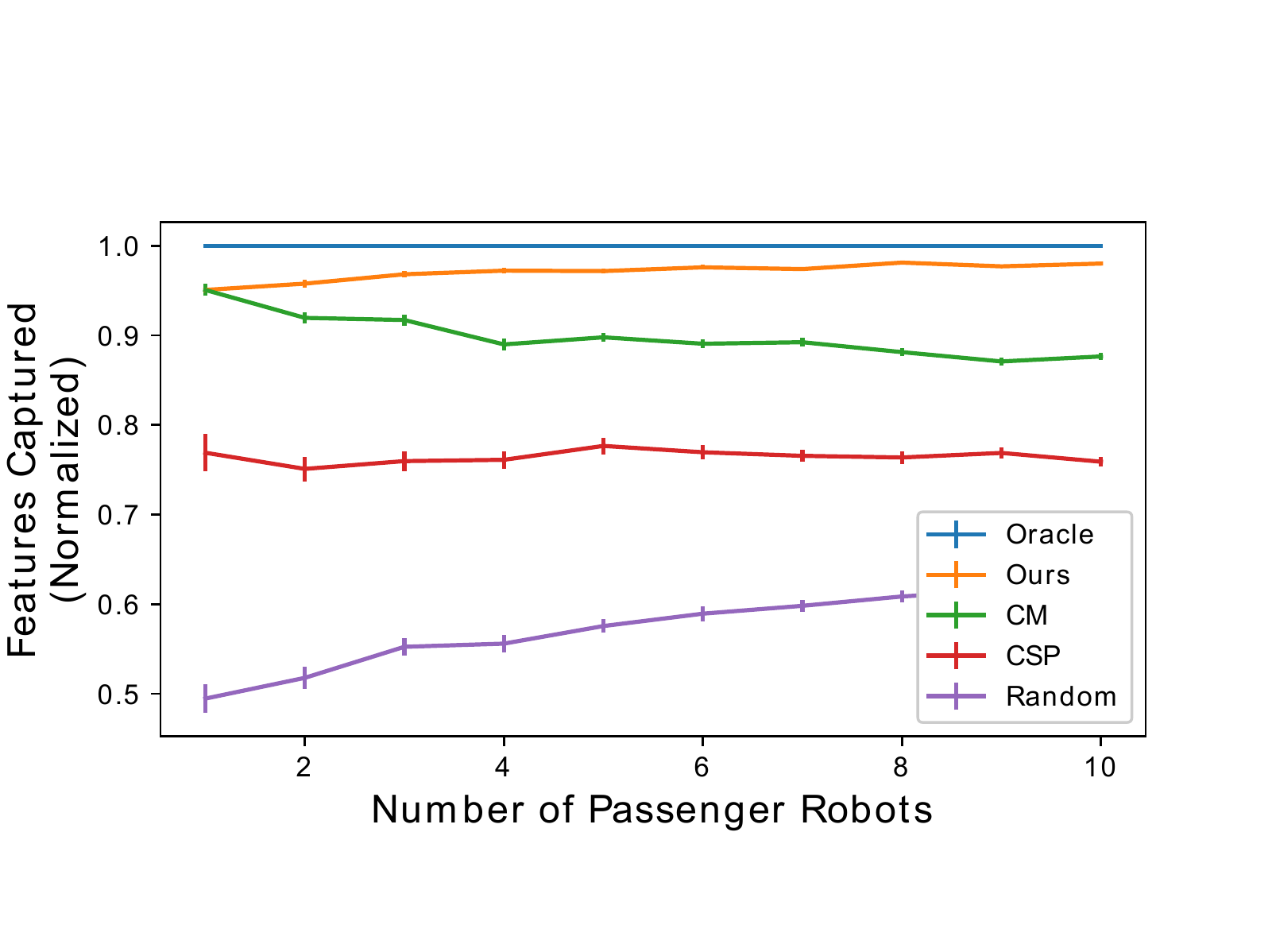}
    \caption{Algorithm comparisons with average utility, as a percentage of the \textit{Oracle}, over $R$ passenger robots and 150 trials. SEM error bars are displayed.}
    \vspace{-4ex}
    \label{fig:multi_passengers}
\end{figure}

Lastly, we show the scalability of \textit{SSAP} to $R$ passenger robots in Fig.~\ref{fig:multi_passengers}. The offline computation of the thresholds is linear in $N$ stages and $R$ number of robots. In practice, the computation of the thresholds takes on the order of seconds. As $R$ increases, the performance of \textit{SSAP} clearly outperforms other algorithms.
\textit{SSAP} is mathematically equivalent to \textit{Cayley Moser} in the case where $R=1$ and yields identical results, as discussed in Sec.~\ref{sec:multi-robot-deployment}. 


\subsection{Aerial Robot Deployment for Exploration}

\subsubsection{Experimental Setup}
\label{sec:real-world-setup}
Recorded LiDAR data from the Urban Circuit of the DARPA Subterranean Challenge~\cite{allen_chung} provided a real-world scenario to test the efficacy of our passenger deployment algorithm. Multiple drone-carrying ground robots autonomously explored the Satsop Nuclear Power Plant in Elma, Washington. For many teams in the competition, a drone deployment was manually triggered by the operator. Identifying an ideal deployment location required constant attention from the operator, already burdened with other tasks. Furthermore, it would be advantageous to utilize prior knowledge from the first robot entering the environment or information from a similar environment.

\begin{figure}[tp]
    \centering
    \includegraphics[width=\linewidth, trim=0mm 2mm 0mm 6mm, clip]{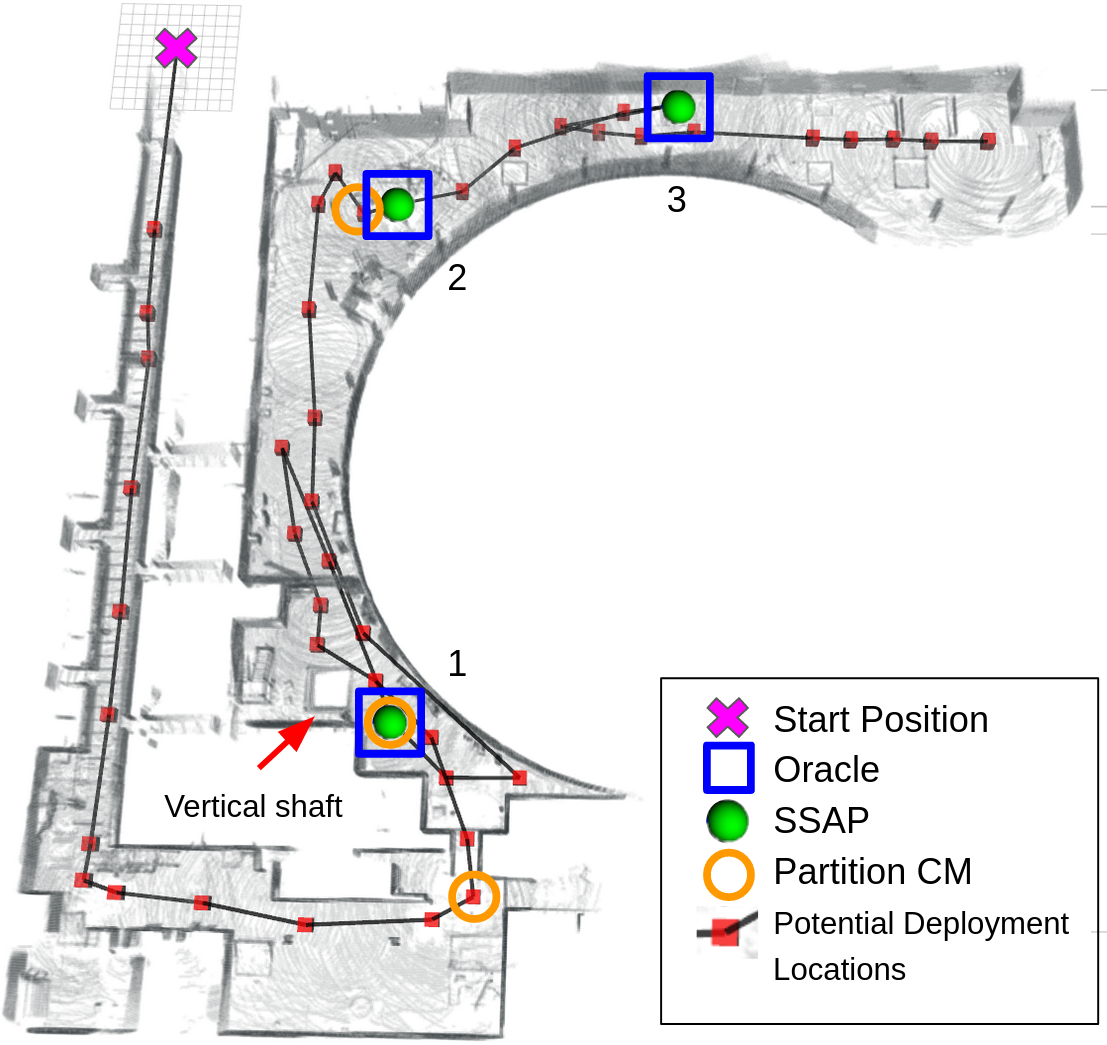}
    \caption{Deployment locations along a 105~m path in the Alpha environment from the DARPA SubT Urban Circuit. The \textit{SSAP} algorithm decides to deploy in the optimal locations. Deploy location 1 is adjacent to a vertical shaft, reachable only by an aerial drone. Location 2 has a higher ceiling and is a larger space than the first \textit{CM} location. Lastly, location 3 leads to the rest of the environment and provides high exploration value.
    }
    \vspace{-2ex}
    \label{fig:real-world-beta-course}
\end{figure}

An occupancy grid of the world was generated from the collected LiDAR data using OpenVDB~\cite{museth2013openvdb}. Using these data, a deployment decision location was set every 2.5 meters along the path that the robot travelled throughout the environment. We defined the value of a deployment location based on the total number of frontier cells within a 10~m radius of the ground robot's location. Frontier cells in the occupancy grid are defined to be free cells neighboring at least one unknown cell~\cite{yamauchi1997frontier}. Furthermore, we filter out frontier cells that are considered accessible to the ground robot. Only frontier cells that are above 1~m and below 0.1~m of the ground robot's position are included. These filtered frontier cells may represent openings such as shafts and ledges that are inaccessible to the ground robots but are ideal for aerial drones. 

Different distributions, representing different prior knowledge, for $f(x)$ in~\eqref{eqn:drl_threshold_recurrence2} were defined in the calculation of the \textit{SSAP} thresholds. In addition to the Poisson distribution, we compared two other distributions utilized by \textit{SSAP}: (1) the histogram distribution and (2) Conway-Maxwell-Poisson (CMP) distribution. First, a distribution was generated from a histogram of the deployment values encountered during the robot's run and can be considered the ground truth distribution for that specific robot's path. Secondly, the CMP distribution generalizes the Poisson distribution and can better handle over/under-dispersion~\cite{shmueli2005useful}, with the pmf
\begin{align}
    \label{eqn:cmp-poisson}
    f(x) &= \frac{\lambda^{x}}{(x!)^\nu}\frac{1}{Z(\lambda,\nu)}, \text{for } x \in \{0,1,...\}
\end{align}
where $Z(\lambda,\nu)$ is a normalization constant. The parameters $\nu$ and $\lambda$ were estimated using a Mean-Squared-Error fit to the histogram data. Similar to the Poisson distribution in Sec.~\ref{sec:simulated_data_setup}, the CMP distribution in~\eqref{eqn:cmp-poisson} can be substituted into~\eqref{eqn:drl_threshold_recurrence2} to generate the \textit{SSAP} thresholds.

\subsubsection{Results}
\begin{figure}[t]
    \centering
    \includegraphics[width=\linewidth, trim=0mm 16mm 0mm 8mm, clip]{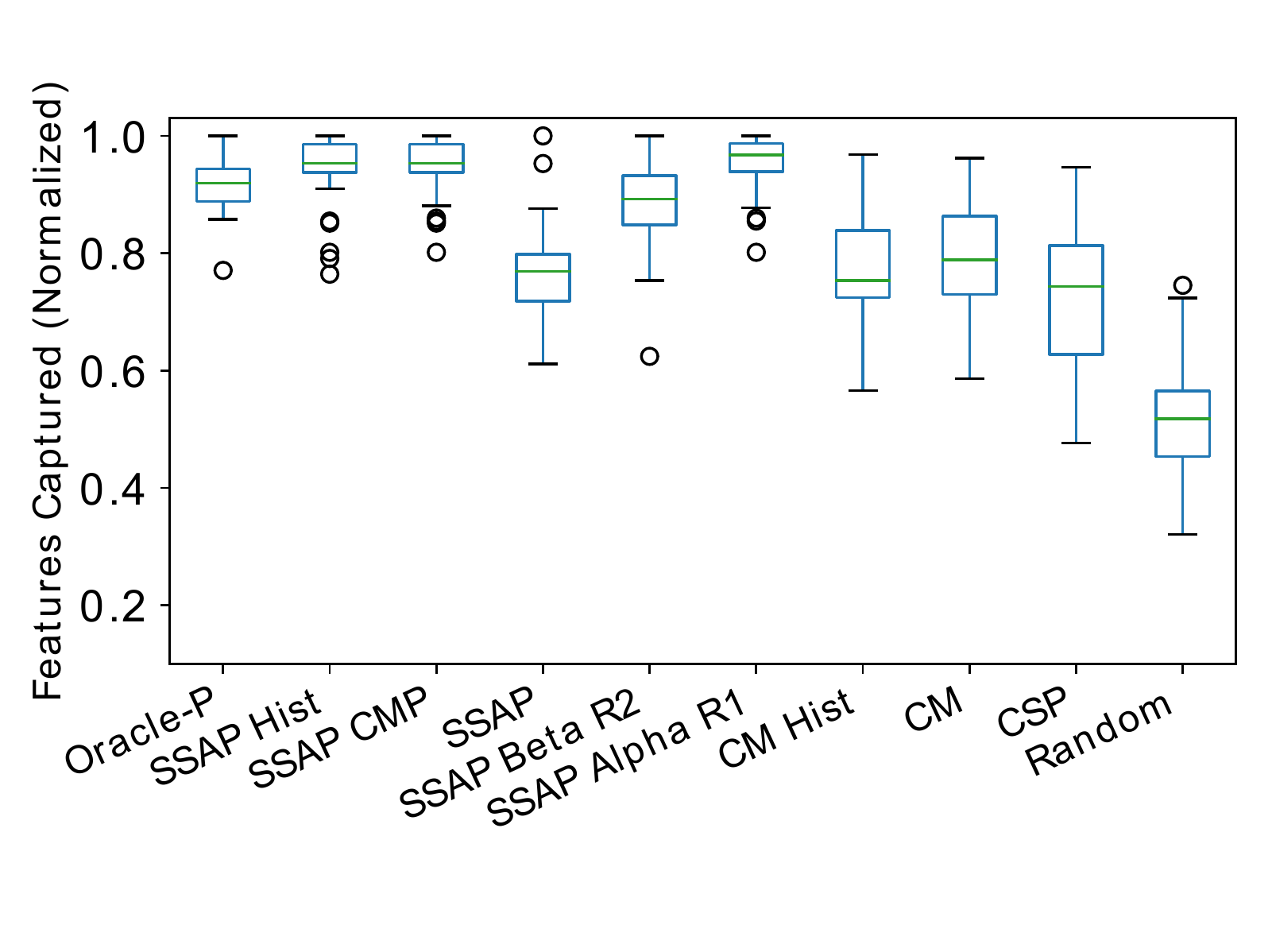}
    \caption{Algorithm comparison results from real data of a robot (R2) exploring Alpha environment. The \textit{SSAP Beta R2} algorithm utilizes histogram data as the prior distribution from the same robot in another environment (Beta). The \textit{SSAP Alpha R1} algorithm uses the histogram data from another robot (R1) in the same environment.}
    \vspace{-2ex}
    \label{fig:real_data_results}
\end{figure}

The \textit{SSAP} algorithm performed strongly in the real-world experiments and outperformed the competing deployment algorithms. The CMP distribution was able to better model the higher variability of real-world deployment values than the Poisson distribution. As seen in Fig.~\ref{fig:real_data_results}, \textit{SSAP} using the CMP distribution performs within 87\% of the oracle. While slightly less than the performance of the ground truth histogram distribution (91\% of oracle), the \textit{SSAP-CMP} distribution outperforms the \textit{SSAP-Poisson} distribution by 18\%. 
For the remaining experiment variations, \textit{SSAP-CMP} is comparable to the \textit{SSAP-Hist} algorithm and outperforms \textit{SSAP-Poisson} by at least 10\% and in one case, up to 28\%. 

We explored two cases of inter-robot information transfer to take advantage of prior domain knowledge: one where the histogram from one robot is used by another robot in the same environment and another where the data from another environment are used. The results of the same environment, inter-robot histogram transfer (\textit{Alpha R1}) can be seen in Fig.~\ref{fig:real_data_results}. The \textit{SSAP} algorithm using the histogram distribution from another robot performs comparably to the high-performing CMP and ground truth histogram distribution, above 90\% of the oracle on average. In the other case, the histogram distribution from a different environment performed on average around 80\% of oracle. Despite a lower performance than inter-robot histogram transfer, the inter-environment histogram transfer (\textit{Beta R2}) can still provide better results than \textit{SSAP-Poisson} or the other baseline methods, \textit{partition Cayley--Moser} and \textit{partition CSP}.




\section{Future Work}
We have provided a formulation of passenger robot deployment for marsupial robots and demonstrated the feasibility of the sequential assignment optimal policy to passenger robot deployments. 
Our results demonstrated that our algorithm enables a significantly improved selection of aerial robot deployment locations for exploration tasks, in comparison to a variety of alternative methods.
In the future, it would be interesting to improve current distribution estimation~\cite{lee2011sequential}, study cases where the prior feature distribution~\cite{albright1977bayesian} or number of stages are not known, and consider a dependent relationship between consecutive observations. Lastly, we aim to extend the deployment problem formulation to multiple carrier robots as well.

\balance


\section*{Acknowledgment}
We would like to thank Rohit Garg for help with the frontier extraction implementation and the SubT Team Explorer~\cite{teamExplorer} for inspiration and providing real-robot data.


\bibliographystyle{IEEEtran}
\bibliography{main}
\end{document}